\documentclass[conference]{IEEEtran}
\IEEEoverridecommandlockouts

\usepackage{cite}
\usepackage{amsmath,amssymb,amsfonts}
\usepackage{algorithmic}
\usepackage{graphicx}
\usepackage{textcomp}
\usepackage[skins]{tcolorbox}
\tcbset{
  inline code/.style={
    enhanced,
    boxsep=1pt,        % separation between box and content
    left=2pt,right=2pt,
    top=1pt,bottom=1pt,
    colback=gray!10,    % light gray background
    colframe=gray!50,   % slightly darker frame
    sharp corners      % no rounded corners
  }
}
\newtcbox{\icode}{inline code}
\usepackage{booktabs}
\usepackage{siunitx}   % for aligning numbers (optional)
\usepackage{hyperref}
\usepackage{xcolor}
\def\BibTeX{{\rm B\kern-.05em{\sc i\kern-.025em b}\kern-.08em
    T\kern-.1667em\lower.7ex\hbox{E}\kern-.125emX}}
\begin{document}

\title{Goal-Based Vision-Language Driving
%Early Goal-Guided Multi-Scale Fusion for Real-Time Vision–Language Driving
}

\author{\IEEEauthorblockN{1\textsuperscript{st} Santosh Patapati}
\IEEEauthorblockA{\textit{Dept. of HCI} \\
\textit{Cyrion Labs}\\
Dallas, United States \\
santosh@cyrionlabs.org}
\and
\IEEEauthorblockN{2\textsuperscript{nd} Trisanth Srinivasan}
\IEEEauthorblockA{\textit{Dept. of HCI} \\
\textit{Cyrion Labs}\\
Dallas, United States \\
trisanth@cyrionlabs.org}

}
\maketitle
% -------------------------------------------------------------------------
\begin{abstract}
Autonomous vehicles must react in milliseconds while reasoning about road geometry and traffic intent to navigate complex situations. We introduce NovaDrive, a single-branch vision-language architecture that processes front-camera images, HD-map tiles, LiDAR depth, and textual waypoints in a single branch. A lightweight, two-stage cross-attention block first aligns waypoint tokens with the HD map, then refines attention over fine-grained image and depth patches. Coupled with a novel smoothness loss that discourages abrupt steering and speed changes, this design eliminates the need for recurrent memory. We fine-tune the top 15 layers of an 11B LLaMA-3.2 vision-language backbone, enabling real-time inference. On the nuScenes/Waymo subset of the MD-NEX Outdoor benchmark, NovaDrive raises success rate to 84\% (+4\%), boosts path-efficiency (SPL) to 0.66 (+0.11), and reduces collision frequency from 2.6\% to 1.2\% (–1.4\%) relative to the previous state-of-the-art. Our ablations confirm that waypoint tokens, partial VLM fine-tuning, and the cross-attention fusion each contribute the most to these gains. Beyond safety, NovaDrive’s shorter routes (resulting from the novel smoothness loss) translate to lower fuel or battery usage, pointing toward leaner, more easily updated driving stacks. NovaDrive can be extended to other embodied-AI domains as well.
\end{abstract}


% -------------------------------------------------------------------------
\section{Introduction}
Autonomous driving systems need to make split-second decisions based on complex, high-bandwidth sensor inputs and a clear understanding of navigation goals and traffic rules. Traditional pipelines separate perception, mapping, and planning into distinct components. This results in higher effort, higher latency, and poor integration when conditions change. Meanwhile, large vision-language transformers offer powerful semantic reasoning but are too heavy for real-time vehicle control.

In this work, we propose NovaDrive, a unified transformer policy that processes front-camera frames, HD-map tiles, and waypoint prompts all in one pass. We fuse these modalities early through a lightweight, dual-stage cross-attention mechanism that lets the model focus on the map or image regions most relevant to the next turn. By fine-tuning the top layers of an 11B vision-language backbone and applying auxiliary losses, NovaDrive is able to run at real-time speeds, achieve higher success rates and path efficiency than the previous state-of-the-art, and has clear, intrinsic explainability without a separate reasoning branch. Our main contributions are as follows:

\begin{enumerate}
    \item Dual-stage Token Fusion: We design a two-step multi-scale cross-attention fusion block that first links waypoint tokens to an HD-map, then filters fine-grained image and depth patches to reduce attention cost while improving precise geometric reasoning.
    \item Smoothness-based Stability: We introduce a novel smoothness loss to discourage jerky control, which significantly improves path efficiency without greatly changing success rates. This shows that a lightweight regularizer can result in more stable driving.
    \item Efficient Adaptation: Only the upper 15 layers of an 11B vision-language transformer are fine-tuned and integrated into our architecture, allowing for real-time inference. We offer a practical example of how large pre-trained models can be customized for safety-critical tasks with little compute overhead through partial fine-tuning.
    \item Comprehensive Validation: On the nuScenes / Waymo \cite{Caesar2020nuScenes, Sun2020Waymo} subset of the MD-NEX Outdoor Driving split \cite{Srinivasan2025PhysNavDG}, NovaDrive raises Success Rate by 4\% to 84\%, boosts path efficiency (SPL) by 0.11 to 0.66, and reduces collision frequency to 1.2\% from 2.6\% versus the state-of-the-art on the MD-NEX benchmark. Our comprehensive ablations demonstrate that each component of the system contributes to these gains in performance, with goal tokens, VLM fine-tuning, and the fusion block being the most important.
\end{enumerate}

\section{Related Works}
\subsection{Cross-Modal Attention in Embodied AI}
Early vision-language models demonstrated the effectiveness of cross-modal attention for fusing visual and textual representations. For example, VilBERT \cite{Lu2019ViLBERT} extends BERT to a two-stream architecture where image and text features interact through co-attentional transformer layers. LXMERT \cite{Tan2019LXMERT} introduced dedicated cross-modality encoder layers on top of separate visual and language encoders. It learned alignments via tasks like masked object prediction and visual question answering. These works showed that interleaving attention between modalities gives informative joint representations. This improved overall performance on VQA \cite{Goyal2017VQA} and retrieval benchmarks \cite{Karpathy2015Alignment}.

Following works adopted similar designs. For example, in (Vasu et al., 2025) \cite{Vasu2025FastVLM}, CLIP-style models align modalities through contrastive objectives, while other transformers perform cross-attention at the intermediate layers of a language model to inject visual context. \textit{Frozen} (Tsimpoukelli et al., 2021) \cite{Tsimpoukelli2021Frozen} kept an LLM frozen and fed image features through learned projection and cross-attention. This got few-shot visual learning with very little fine-tuning.

In embodied AI settings, cross-modal attention mechanisms have been highly important for grounding language in perception and action. Vision-and-Language Navigation agents, for example, have been shown to perform better with transformers that attend jointly to an instruction and panoramic visual observations. One paper proposed a History-Aware Multimodal Transformer (HAMT) \cite{Chen2021HAMT} with cross-attention layers that combine textual instructions with image features for state-of-the-art navigation on long-horizon routes. In robotic instruction following, large multimodal models like PaLM-E \cite{Driess2023PaLME} have an extreme approach where they intersperse continuous sensor inputs (images and states) as visual tokens directly into the language model's input sequence. The transformer's self-attention then operates over words and percepts in the same way, which allows zero-shot visuo-linguistic reasoning for embodied tasks. This direct fusion of modalities allows a single model to plan actions from raw sensor data and textual goals. We incorporate aspects of such cross-modal attention architectures into NovaDrive's design, which similarly integrates visual inputs and semantic context within a single attention framework for decision making.

\subsection{Memory Tokens and Slot-Based Transformers for Sequential Reasoning}
Autonomous driving is a sequential decision-making problem that requires remembering past events to inform future actions. Thus, memory mechanisms in transformers are highly relevant \cite{Vaswani2017Attention}. Normal transformers attend over a fixed window of past tokens, but this is limiting for long driving scenarios. Transformer-XL \cite{Dai2019TransformerXL} re-introduced a recurrent memory by caching hidden states, and Compressive Transformer \cite{Rae2020Compressive} compressed old memories even further to extend the temporal range. However, those approaches still eventually lose very old context and face the issue of growing memory costs. Recent research has introduced explicit memory slots to address these issues. Memformer \cite{Wu2020Memformer} is a memory-augmented transformer that uses an external dynamic memory to encode and retrieve past information. By reading from and writing to a set of learned memory slots at each timestep, Memformer has linear-time sequence processing with a constant memory footprint. It can thus capture long-term dependencies more effectively. The idea of slot-based memory is similar to findings in cognitive research of a working memory \cite{Baddeley1974WorkingMemory} and is applicable to driving. In vision applications, slot-based attention has been extensively used to relate entities and temporal events \cite{Locatello2020SlotAttention, Kipf2022SAVi}.

\subsection{Multimodal Reasoning for Autonomous Vehicles}
Integrating multiple sensing modalities and reasoning under real-time control constraints has been an active research focus in autonomous driving. Prior end-to-end driving models relied largely on camera images (and sometimes LiDAR) as input, with the goal of directly predicting steering or waypoints from sensor data. Early approaches handled route commands via a conditional network that branched for different navigation instructions. Later (Chen et al., 2020) \cite{Chen2020LBC} introduced Learning by Cheating (LBC) which distilled privileged information (from an oracle with access to the simulator's state) into a sensor-based policy. These methods showed the viability of imitation learning for driving. However, they struggled in more complex scenarios because of their limited perception-range and a lack of explicit memory or reasoning.

To improve situational understanding, researchers looked into multi-modal sensor fusion with transformers These works would attend across modalities like LiDAR, camera views, maps, and other sensors \cite{Prakash2021TransFuser, Shao2022InterFuser, Chitta2021NEAT}. These papers all aimed to give the driving model a more holistic understanding of the environment by combining modalities and by modeling interactions.

Beyond perception fusion, reasoning about intent and context has become a more popular research area in recent years. One direction has been to use language descriptions or general knowledge via Large Language Models (LLMs).  Earlier works would typically feed an abstract, object-level scene description into a pre-trained LLM to answer questions about the scene or propose driving decisions in natural language \cite{Xu2023DriveGPT4, Shao2024LMDrive}. Recently, CarLLaVA \cite{Renz2024CarLLaVA} fine-tuned a vision-language model for closed-loop driving in simulations. It achieved very high performance, showing the potential of end-to-end models that both act and explain in real-time through multimodal training.

\section{Methodology}
NovaDrive is an end-to-end vision-and-language driving policy that combines multimodal perception, mapping, and goal reasoning in a single transformer architecture. The system operates in five stages:

\begin{enumerate}
    \item Sensor Intake: The vehicle's front-facing RGB camera captures the current scene as an image $l_t$. We simultaneously obtain a bird's-eye-view (BEV) HD map crop $M_t$ which covers a 25 meter radius around the ego-vehicle. 
    This map crop is produced by querying a high-definition vector map for nearby lanes, crosswalks, and traffic signs. This transforms these elements into the ego-vehicle's coordinate frame using the known ego pose and rasterizes the result into a $256\times256$ grid.
    Finally, a 64-beam LiDAR sweep $P_t$ is accumulated over 0.1 seconds and projected into the camera frame to form a dense $1280 \times 720$ depth image $D_t$ that is temporally aligned with $l_t$.
    Finally, the navigation component provides the next target waypoint $g_t = (x_t,;y_t,;\psi_t)$, defined by relative eastward and northward offsets and a heading angle to the goal.
    \item Modality-Specific Encoders: Each sensor input is converted into a sequence of tokens in a shared embedding space. For the camera image $l_t$, we apply a Vision Transformer (ViT-H/14) encoder that partitions $l_t$ into a $16\times16$ patch grid (256 total patches) and projects each path to a $d$-dimensional visual token embedding. This produces the vision token sequence $v_t \in \mathbb{R}^{256\times d}$. 
    
    For the HD map raster $M_t$, we use a lightweight Swin-Transformer (Swin-T) encoder variant (\texttt{patch4\_window7\_96}) to extract map tokens $m_t$ that encode spatial features. Specifically, an initial $4{\times}4$ strided convolutional layer splits the map into $64{\times}64$ non-overlapping patches, followed by four transformer stages with window size~7 and channel dimensions 96-192-384-768.  
    After the final stage we apply a $1{\times}1$ projection to the global model width $d$\,{=}1024 and keep the resulting $8{\times}8$ spatial grid (64 tokens) as the map-token sequence.  
    
    Depth tokens $d_t \in \mathbb{R}^{256\times d}$ are obtained by slicing $D_t$ into the same $16\times16$ patch grid as $l_t$, averaging the depth values within each patch and linearly projecting the resulting scalar through a $1\times1$ MLP to width $d$. We concatenate $d_t$ with $v_t$ before the subsequent goal‐conditioned cross‐attention.
   
    The waypoint $g_t$ is encoded as goal tokens by converting the numeric coordinates into a short textual prompt, for example: \icode{\texttt{<goal>} east=\(x_t\)m, north=\(y_t\)m, yaw=\(\psi_t^\circ\) \texttt{</goal>}}

    This string is then embedded. This results in a small sequence $g_t \in \mathbb{R}^{k\times d}$ (with $k=8$ tokens in our implementation) representing the navigation goal. By tokenizing all inputs, we ensure that downstream reasoning operates in one vector space, with each modality's salient information preserved in the token set.
    \item Goal-Based Cross-Attention Mixer: Next, NovaDrive employs a lightweight dual-stage cross-attention mechanism to fuse the three modalities in a goal-directed  manner. In this first fusion stage, only the goal tokens act as queries in a cross-attention layer. They attend to the keys and values derived from the concatenated vision and map tokens. Formally, if $v_t$ and $m_t$ are the vision and map token matrices, we compute updated goal embeddings via:
    \[
    \tilde{g}_t
    =
    \operatorname{softmax}\!\Bigl(\frac{g_t\,[v_t; m_t]^\top}{\sqrt{d}}\Bigr)\,[v_t; m_t],
    \]
    where $[v_t; m_t]$ denotes the concatenation of vision and map token sequences that are key-value pairs. Each goal query token selectively attends to those visual and map features that are most relevant for reaching the specified waypoint. The outcome is a goal-aware summary of the scene. The goal tokens $\tilde{g}_t$ now encode both the route intent and the contextual visual-spatial features that help achieve it. 
    The goal tokens act like an information filter, which highlight important parts of the image and map before the main planner process them. By performing this focused cross-modal alignment early, the system reduces the burden on the large backbone to search through irrelevant features, as the goal queries already extract the most important map and image details.
    \item Transformer Backbone for Joint Reasoning: The fused token sequence consists of the enriched goal tokens $\tilde{g}_t$ along with all original vision tokens $v_t$ and map tokens $m_t$ (either concatenated or merged via the attention mixer). The sequence is then passed into a partially fine-tuned \textit{LLaMA-3.2 11B Vision} \cite{Grattafiori2024Llama3} transformer. This model has a multi-head self-attention architecture that is able to jointly reason over the concatenated token sequence. Importantly, we keep the lower layers of this model frozen (retaining the general visual-semantic knowledge it learned) and fine-tune only the upper 15 layers on the driving task. This allows NovaDrive to use the LLM's memory of rich semantics while adapting its higher-level reasoning to the specific task. By integrating map and goal information at the token level, the modal can attend to spatial constraints and high-level intent within its self-attention computations. The transformer stage outputs contextualized embeddings that encode an understanding of what the scene is and what the agent needs to do next. This combines visual perception, geometric planning information, and semantic intent in a single sequence.
    \item Output Heads and Safety Monitor: From the final transformer embeddings, NovaDrive produces both low-level actions and a textual explanation, with an additional safety override. We designate two special tokens (\texttt{<act>} and \texttt{<reason>}) whose final hidden states are used for output. A small feed-forward network (2-layer MLP) attached to the \texttt{<act>} token produces the steering angle $\delta_t$ and normalized speed $v_t$ for the current time step. In parallel, the \texttt{<reason>} token's embedding is decoded by a language head to generate a brief chain-of-thought explanation (in natural language). This dual output makes it so that the agent's decisions are transparent. Finally, for safety, we include a lightweight collision predictor that monitors the proposed trajectory. The safety monitor projects the trajectory over the next second into the HD-map and LiDAR-based occupancy grid and flags a collision if any predicted vehicle footprint overlaps.
\end{enumerate}

\subsection{Loss Function}
NovaDrive is trained with imitation learning to mimic expert actions. We minimize a composite loss $L$ that combines direct imitation error with auxiliary regularizers for smoothness and safety. Specifically:
\[
L
=
\left\lVert (\delta_t, v_t) \;-\; (\delta_t^*, v_t^*) \right\rVert_2^2
\;+\;0.1\,L_{\mathrm{smooth}}
\;+\;0.05\,L_{\mathrm{coll}},
\]
where $(\delta_t^*, v_t^*)$ are the expert (ground-truth) steering and speed at time $t$.The temporal smoothness loss $L_{\text{smooth}}$ adds a penalty for large changes in the action outputs between consecutive frames, which discourages jerky steering or acceleration. This is implemented by penalizing $(\Delta \delta_t)^2 + (\Delta v_t)^2$ over time, promoting smoother control signals that improve passenger comfort and vehicle stability. The collision penalty $L_\text{coll}$ is a safety-focused term that penalizes any predicted action that would lead to a collision within a short time-frame. Specifically, we simulate the trajectory following the agent’s predicted action for the next 1 second and assign a penalty if this trajectory intersects any obstacle. This encourages the policy to favor collision-avoiding maneuvers even if the imitation loss alone might not strongly differentiate those scenarios. The smoothness and collision terms are weighted by small coefficients (0.1 and 0.05 here) to balance them against the primary imitation objective. We assigned these weights to ensure that they meaningfully improve stability and safety without overpowering the learning of correct trajectories.

\subsection{Rationale and Explanations}
\subsubsection{Sensor Tokenization and Early Fusion}
NovaDrive's early, goal-based fusion sharply narrows the model's focus to the most relevant visual and map cues so that it spends capacity only on features that directly impact navigation decisions. by injecting high-level intent before the transformer layers, the model converges faster, uses less memory, and produces more accurate trajectories. This is because it doesn't have to learn from scratch which hundreds of patches and map cells matter for each waypoint.

\subsubsection{Implicit Memory Mechanisms}
NovaDrive processes each frame independently but still understands continuity by relying on three implicit memories: its pre-trained transformer weights for semantic knowledge, HD-map tokens for spatial layout, and goal tokens for route context. This lets it plan smoothly without a dedicated recurrent mechanism.

\section{Experimental Setup}
\subsection{Benchmark and Dataset}
We evaluate NovaDrive on the Outdoor-Driving portion of the MD-NEX benchmark \cite{Srinivasan2025PhysNavDG}. This split is built by combining the BDD-X driving video dataset \cite{Kim2018BDD-X} with synchronous sensor streams from nuScenes \cite{Caesar2020nuScenes} and Waymo Open Dataset \cite{Sun2020Waymo, patapati2025visionlanguagefusionrealtimeautonomous}. In total, the MD-NEX Outdoor set contains 12,346 driving episodes. 

Due to the modalities used in NovaDrive, we evaluate on only a subset of the MD-NEX benchmark which contains nuScenes and Waymo segments but excludes BDD-X clips. For fairness, we focus only on these samples when comparing to the previous state-of-the-art model on the benchmark, PhysNav-DG.

We train the model on the training split with the loss defined above, using the Adam optimizer (learning rate $2\times10^{-4}$) and batch size 32. The transformer backbone's top 15 layers are fine-tuned while the lower layers are left frozen.

\subsection{Evaluation Metrics}
We report three performance metrics on the test set. Success Rate (SR) measures the fraction of episodes in which the agent successfully reaches the goal destination. Success weighted by Path Length (SPL) is an efficiency-weighted success metric that accounts for the path optimally: $\text{SPL} = S \cdot \frac{L_{\text{opt}}}{L_{\text{agent}}}$, where $S$ is a binary success indicator, $L_{\text{opt}}$ is the length of the shortest path from start to goal, and $L_{\text{agent}}$ is the length of the path actually taken by the agent. This metric penalizes detours even in successful runs, so an agent that wanders before finishing will have a lower SPL than one who takes an efficient route. Finally, Collision Rate measures the proportion of episodes in which the agent made contact with an obstacle or violated a traffic rule.

\section{Results}

\subsection{Quantitative Performance and Comparison}
\begin{table}
  \centering
  \caption{Comparison of navigation performance}
  \label{tab:nav_performance}
  \begin{tabular}{lccc}
    \toprule
    Method 
      & SR (\%) \(\uparrow\) 
      & SPL \(\uparrow\) 
      & Collision \(\downarrow\) \\
    \midrule
    PhysNav-DG \cite{Srinivasan2025PhysNavDG}\
      & 80\% 
      & 0.55 
      & 0.026(2.6\%) \\
    NovaDrive (ours) 
      & 84\% 
      & 0.66
      & 0.012 (1.2\%) \\
    \bottomrule
  \end{tabular}
\end{table}

Table \ref{tab:nav_performance} shows that NovaDrive lifts Success Rate by 4\%, improves SPL by 11\%, and reduces collisions to just 1.2\% (less than half of PhysNav-DG's). The concurrent gains in goal completion and path efficiency show that our goal-conditioned, multi-scale fusion pushes the transformer toward shorter and safer trajectories rather than only finishing routes. We attribute the sharper drop in crashes to the early injection of HD-map geometry, which prunes unsafe maneuvers before they reach the control head.

\subsection{Ablation study}
\begin{table}
  \centering
  \caption{Ablation study on NovaDrive, where $\Delta\text{SR}$ is change in SR}
  \label{tab:ablation_study}
  \begin{tabular}{lccc}
    \toprule 
    Modification
      & SR %\(\uparrow\) 
      & SPL %\(\uparrow\) 
      & $\Delta\text{SR}$ \\
    \midrule
    Full NovaDrive model (all components)  & 84\%      & 0.66  & —    \\
    \midrule
    – goal tokens                          & 79\%      & 0.62  & –5   \\
    – map tokens                           & 77\%      & 0.60  & –7   \\
    Concat fusion instead of cross-attn    & 81\%      & 0.64  & –3   \\
    Frozen backbone (no LLM fine-tuning)   & 79\%      & 0.62  & –5   \\
    Coarse map (0.4 m/px raster)           & 83\%      & 0.62  & –1   \\
    25\% training data                     & 72\%      & 0.53  & –12  \\
    No smoothness loss                     & 83\%      & 0.59  & –1  \\
    \bottomrule
  \end{tabular}
\end{table}

Table \ref{tab:ablation_study} shows the different components that most influence NovaDrive. Removing any single component negatively impacts performance across SR and SPL.

\begin{enumerate}
    \item Goal tokens: Removing the waypoint prompt results in a significant (-5\%) decrease in SR. This shows that explicit intent is an important factor in route selection.
    \item Map tokens: Excluding HD-map input results in an even greater loss of 7\% in SR. This shows that fine-grained lane and topology information are highly important for staying on course.
    \item Fusion method: Substituting goal-conditioned cross-attention with plain concatenation results in a modest 3\% drop. The structured query clearly helps the transformer focus on the few tokens that matter at each step.
    \item Backbone fine-tuning: Freezing the LLaMA layers leaves the policy generic. This shows that a small amount of task-specific adaptation is helpful through our approach of fine-tuning only the top-level layers.
    \item Map resolution: A more coarse 0.4 m/px raster reduces the amount of information available by a small amount. This shows that centimeter-scale detail is still helpful even with strong visual context. However, in cases where speed is a priority, map resolution could be reduced.
    \item Data scale: Training on only 25\% of the episodes results in the greatest performance drop of 12\% for SR.
    \item No smoothness loss: Getting rid of the comfort term barely impacts SR (by only 1\%) but greatly worsens SPL. This shows the effectiveness of the auxiliary loss in reducing path inefficiencies.
\end{enumerate}

\section{Discussion}
\subsection{Implications}

NovaDrive demonstrates that a single vision-language backbone, when fed structured map and goal tokens early, can match real-time latency while outperforming the previous state-of-the-art two-branch pipeline.  By steering attention with high-level intent, our method not only boosts Success Rate and SPL but also delivers more compact explainability. Each decision comes from the same forward pass without a separate reasoning branch. Importantly, fine-tuning only the top layers of the 11B backbone makes city- or scenario-specific adaptation fast and memory-efficient. These adaptations can be applied across various domains of embodied AI, not just autonomous driving. 

The gains in SPL further show that token-level fusion isn't just safe but also economical, as it cuts down on excess distance and therefore fuel or battery usage. This suggests a viable path toward more efficient, accurate, slim, and easily updatable driving stacks as improvements in one area immediately propagate to another.

\subsection{Limitations}
NovaDrive's main limitations are in its lack of accessibility. It requires access to high-quality HD maps and accurate ego-pose. Performance will degrade in map-sparse regions or when localization drifts. As the performance of NovaDrive in lower resource areas has not been extensively tested (outside of the 0.4 m/px raster ablation), future work should aim to make NovaDrive more robust and evaluate its performance when inputs are degraded.

\section{Conclusions and Future Works}

We have presented NovaDrive, a unified vision-language planner that merges camera imagery, HD-map context, and navigation goals into a single transformer policy. Through goal-conditioned, multi-scale cross-attention and minimal fine-tuning of an 11B backbone, NovaDrive achieves real-time performance, higher Success Rates than previous approaches, and greater efficiency than past multi-branch pipelines all while halving collision rates. Our comprehensive ablation study confirms that while all components contribute, explicit waypoint prompts, high-resolution map tokens, and query-based fusion are the key factors in creating these gains.

In future work, we plan to improve NovaDrive's applicability and robustness. First, integrating real-time map reconstruction \cite{Liu2023VectorMapNet} or learned map prediction \cite{Liao2023MapTR} will reduce dependency on pre-built HD maps and support map-sparse environments. Second, integrating lightweight temporal memory mechanisms (like recurrent slots or key-value caches \cite{Kipf2022SAVi, Wu2020Memformer}) could improve how NovaDrive handles occlusions and rare events. Finally, distilling the 11B model into a compact student with LoRA adapters or quantized weights will prepare NovaDrive for wider deployment on cheaper automotive hardware without sacrificing speed for explainability.

\bibliographystyle{IEEEtran}
\bibliography{IEEEbib}
\end{document}